\documentclass{article}

\usepackage{microtype}
\usepackage{graphicx}
\usepackage{booktabs} % for professional tables

\usepackage{hyperref}
\usepackage{url}
\usepackage{mathtools}
\usepackage{amsthm}
\usepackage{bm}
\usepackage{subcaption}
\usepackage{multirow}
\usepackage{graphics}
\usepackage{float}
\usepackage{array}
\usepackage{caption}

\usepackage{algpseudocode}

\usepackage[accepted]{icml2025}

\usepackage{amsmath}
\usepackage{amssymb}
\usepackage{mathtools}
\usepackage{amsthm}
\usepackage[capitalize,noabbrev]{cleveref}
\usepackage{subcaption}

\theoremstyle{plain}
\newtheorem{theorem}{Theorem}[section]
\newtheorem{proposition}[theorem]{Proposition}
\newtheorem{lemma}[theorem]{Lemma}

\theoremstyle{definition}

\theoremstyle{remark}

\usepackage[textsize=tiny]{todonotes}

\begin{document}

\twocolumn[
\icmltitle{Background-Aware Defect Generation for Robust Industrial Anomaly Detection} 



\icmlsetsymbol{equal}{*}

\begin{icmlauthorlist}
\icmlauthor{Youngjae Cho}{yyy}
\icmlauthor{Gwangyeol Kim}{yyy}
\icmlauthor{Sirojbek Safarov}{yyy}
\icmlauthor{Seongdeok Bang}{yyy}
\icmlauthor{Jaewoo Park}{yyy}

\end{icmlauthorlist}

\icmlaffiliation{yyy}{AiV Co.}

\icmlkeywords{Machine Learning, ICML}

\vskip 0.3in

]
% \printAffiliationsAndNotice{}  



\begin{abstract}

% In anomaly detection, the scarcity of anomalous data challenges deep neural networks in identifying anomalies. Generative models can address this imbalance by synthesizing anomaly data, but existing methods often neglect the critical relationship between the defect and its background. Defects depend on the background context, and training only on the defect area can lead to biased or unrealistic generations, especially for logical anomalies where the defect must align with the background (e.g., an orange defect on an orange juice bottle). We propose a method that explicitly models the background-defect relationship, where the background influences defect denoising without affecting the background itself. Our approach introduces a disentanglement loss to separate the denoising processes, enhancing defect generation. We also use DDIM Inversion to generate defects on target normal images while preserving the background. We prove that our method generates realistic defects with invariant backgrounds and demonstrate its effectiveness through experiments.

Detecting anomalies in industrial settings is challenging due to the scarcity of labeled anomalous data. Generative models can mitigate this issue by synthesizing realistic defect samples, but existing approaches often fail to model the crucial interplay between defects and their background. This oversight leads to unrealistic anomalies, especially in scenarios where contextual consistency is essential (i.e., logical anomaly). To address this, we propose a novel background-aware defect generation framework, where the background influences defect denoising without affecting the background itself by ensuring realistic synthesis while preserving structural integrity. Our method leverages a disentanglement loss to separate the background’s denoising process from the defect, enabling controlled defect synthesis through DDIM Inversion. We theoretically demonstrate that our approach maintains background fidelity while generating contextually accurate defects. Extensive experiments on MVTec-AD and MVTec-Loco benchmarks validate our method’s superiority over existing techniques in both defect generation quality and anomaly detection performance.

\end{abstract}

\section{Introduction}

\begin{figure*}
    \includegraphics[width=0.49\linewidth]{figure/generation.pdf}
    \includegraphics[width=0.49\linewidth]{figure/loco1.pdf}
    
    \caption{Left image is a comparison between ours and baselines for MVTec AD. Right image is a comparison between ours and baselines for MVTec Loco.}
    \label{fig:Generation1}

\end{figure*}

\label{sec:intro}

% Data scarcity is a core problem in industrial anomaly detection. Imbalanced data induces overfitting issues in the deep neural network, which makes model prediction inaccurate. Limited labeled train instances induce model bias, which prevents the detection model from generalizing to unseen anomaly instances. Recent unsupervised-based anomaly detection \citep{roth2022towards,deng2022anomaly,liu2023simplenet} avoids the imbalanced issues without fitting to anomaly instances, but it can be a problem when the backbone model undergoes the distribution shifts in the feature space and normal features are not discriminative with anomalous features and it is hard to detect fine-grained anomalous areas at the pixel-level. 

Data scarcity presents a significant challenge in industrial anomaly detection, as imbalanced datasets can lead to overfitting in deep neural networks, damaging inspection performance. Recent unsupervised anomaly detection approaches \citep{roth2022towards,deng2022anomaly,liu2023simplenet} mitigate the imbalance issue by avoiding reliance on anomaly instances during training. However, these methods may struggle when the backbone model experiences distribution shifts in feature space, making it difficult to distinguish normal and anomalous features. Additionally, they often fall short in detecting fine-grained anomalies at the pixel level.

Recent research on defect generation aims to address data scarcity through a data-centric approach, synthesizing anomaly data to train on augmented samples. GAN-based methods \citep{duan2023few} generate realistic synthetic anomaly data from limited examples, improving model performance on rare anomalies; however, they lack control over defect location, often transforming normal regions into anomalies unintentionally. Diffusion-based methods \citep{hu2024anomalydiffusion} improve control by specifying defect locations and blending normal latent background with synthetic defect latent. While these methods have made strides in generating faithful anomaly samples, they fail to adequately model the critical relationship between the defect and its background. 

Defects are closely related to the background, making it difficult to control their generation using only mask information. If a model relies solely on a given mask, the generated defect may appear out of place, failing to integrate naturally with the object’s structure. Even when the defect is generated within the masked region, it can overfit to the mask shape rather than capturing realistic variations, leading to unrealistic anomalies in Figure \ref{fig:Generation1}, where defect is not aligned with the background. Such inconsistencies in generated data can ultimately degrade the robustness of anomaly detection models.

Furthermore, in the case of logical anomalies, defects are not always visually apparent, making it necessary to leverage background information to detect them effectively. Since these anomalies are defined not just by their appearance but by their contextual inconsistency, modeling the relationship between the defect and its background is essential for generating meaningful and realistic anomalies, ultimately enhancing the robustness of the detection model.

While modeling the relationship between defects and their background is crucial, it is essential to structure this relationship so that the background influences the defect rather than the reverse. Achieving controllability in defect generation requires the model to generate defects directly on target normal samples, where defect is not contained. However, training a generative model with anomaly samples presents a challenge: defects and background are inherently entangled in these examples, making it difficult for the model to separate their contributions. By decoupling the background from the defect during training, model can generate realistic defects that align with the characteristics of the background in the target normal sample.

% In real-world anomaly detection, defects are not independent of the background, where they are influenced by the surrounding context, such as texture, color, and spatial positioning. Depending on only mask information, generated defects may appear out of place, disrupting the natural structure of the object and leading to unrealistic anomalies.  %logical anomaly에 대한 설명을 좀 더하기 
% Despite the importance of modeling the relationship between the defect and its background, it is essential that the relationship is structured such that the background influences the defect, rather than the other way around. To achieve controllability in defect generation, the model must be capable of generating defects directly on target normal samples. However, training the generative model using anomaly samples introduces a challenge: the defect and background are intertwined in these training examples, making it difficult for the model to disentangle their contributions. By decoupling the background from the defect during training, the model can generalize better, generating realistic defects that align with the characteristics of the background in the target normal sample.

 Therefore, our work suggests the disentanglement loss function for generating faithful defects on the target normal instances. Our formulation ensures independence from defect to background, allowing the denoising process for defects to proceed without affecting the background's denoising process. At the same time, the background influences the defect area’s denoising process, enhancing defect generation by reducing reliance on the mask information. In addition, we propose the inference strategy to generate defect on the target normal latent, where we apply DDIM Inversion to the normal latent and generate defects on the background.  We theoretically show that we can initialize anomaly latent by noising the target area of inversed normal latent while reconstructing its background. To sum up, we make the following contributions. 
\begin{itemize}
\item We introduce a regularization loss that generates defect based on background but without influencing background generation with defect characteristics. 
% \item We introduce regularization loss for disentanglement from defect to background for modeling background and defect area in anomaly instances\\
% \item We theoretically show that our loss formulation can inverse anomaly latent given background of latent encoded from background image instance via DDIM Inversion.  
\item We theoretically show that our loss formulation can inverse anomaly latent given the latent of background under the DDIM inversion and maintain the background reconstruction. 

\item We demonstrate that our methodology can generate faithful and diverse generations and effectively mitigate data scarcity for detecting structure and logical anomalies. 
\end{itemize}
\section{Background}
\label{sec:Background}

\subsection{Industrial anomaly detection}
Industrial anomaly detection (IAD) has seen significant progress, transitioning from traditional methods reliant on labeled data to more advanced machine learning and computer vision approaches. By leveraging pretrained models to extract normal features, unsupervised anomaly detection methods focus on identifying anomalous regions in products \citep{defard2021padim, cohen2020sub, roth2022towards, deng2022anomaly, liu2023simplenet}. Memory-based approaches \citep{defard2021padim, cohen2020sub, roth2022towards} store normal feature representations and calculate similarity scores to pinpoint anomalous regions. However, since unsupervised methods do not incorporate anomalous features during training, accurately detecting defect areas in anomaly instances remains challenging. As a result, supervised or semi-supervised learning methods remain highly effective for identifying fine-grained defect areas at the pixel level, especially when sufficient anomaly instances are available for training.

% Exploiting pretrained-model to convert the normal features, unsupervised anomaly detection inspects the anomalous area of products \citep{defard2021padim,cohen2020sub,roth2022towards,deng2022anomaly,liu2023simplenet}. Memory-based methods \citep{defard2021padim,cohen2020sub,roth2022towards} save normal feature representation and calculate similarity score to locate anomaly features. Since unsupervised anomaly detection does not exploit the anomalous feature in training, it is hard to detect defect area of anomaly instances precisely. Therefore, supervised or semi-supervised learning methods are still powerful for detecting fine-grained defect areas at the pixel level when sufficient anomaly instances are given during training.   
\subsection{Defect generation}

Collecting sufficient anomalous data is challenging due to the time-consuming nature and high annotation costs involved. To address this scarcity, data augmentation and generation methods have been developed. Cut-Paste \citep{li2021cutpaste} employs pixel-level data augmentation, while Crop-Paste \citep{lin2020few} focuses on feature-level augmentation to help the model learn discriminative features between normal and anomalous data. However, since these augmentation-based methods do not model distributions of anomalies, generating a faithful defect dataset remains difficult.

% Since collecting enough anomalous data is hard (i.e., time-consuming, annotation cost), data augmentation and generation methods are conducted to alleviate this scarcity issue. Cut-Paste \citep{li2021cutpaste} utilizes the pixel-level data augmentation, and Crop-Paste \citep{lin2020few} utilizes the feature-level data augmentation for learning the discriminative features between normal and anomalous data. Since the augmentation-based methods do not learn the distribution of anomalies, it is hard to synthesize the diverse and faithful defect dataset.

In the DFMGAN \citep{duan2023few} paper, the authors leverage Style-GAN \citep{Karras_2020_CVPR} to generate defects from a limited number of anomalous data. While they also incorporate normal data to guide defect generation, their approach lacks the ability to control defect generation on unseen target normal data. In contrast, AnomalyDiffusion \citep{hu2024anomalydiffusion} addresses this by training a Text-to-Image diffusion model to control defect generation on target normal images. However, their method does not train the diffusion process to account for the background of anomalous data, meaning that defect generation does not consider the relationship between the defect and the target normal background. Since some defects are dependent on the background, it is crucial to model the interaction between the background and defect to generate more faithful and contextually accurate defects.

\begin{figure*}[h!]
\centering
     \resizebox{0.9\linewidth}{!}{
    \includegraphics[width=0.48\linewidth]{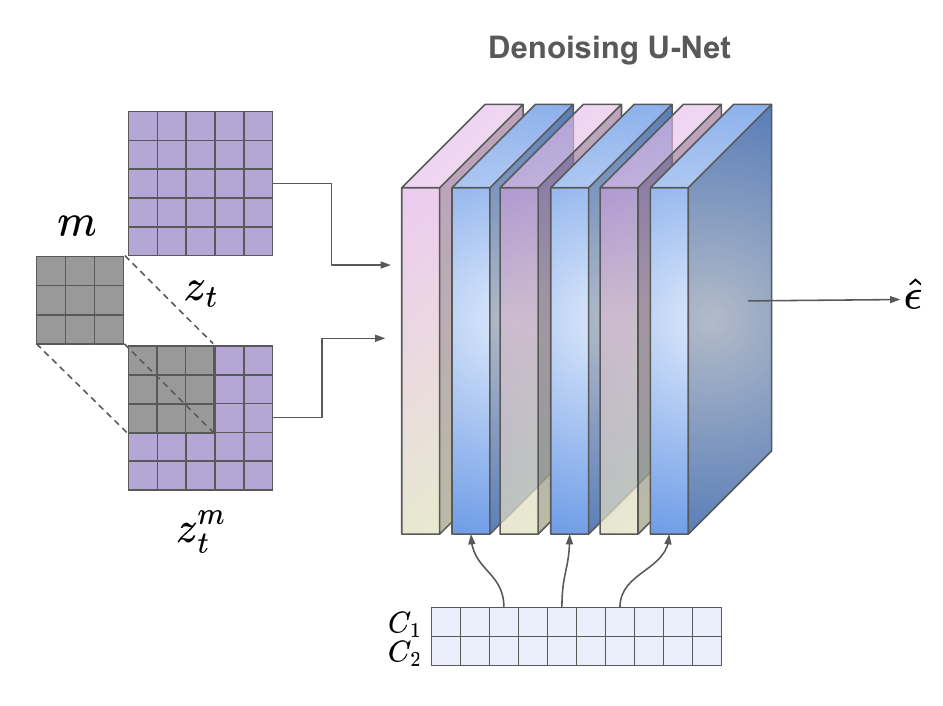}
    \includegraphics[width=0.48\linewidth]{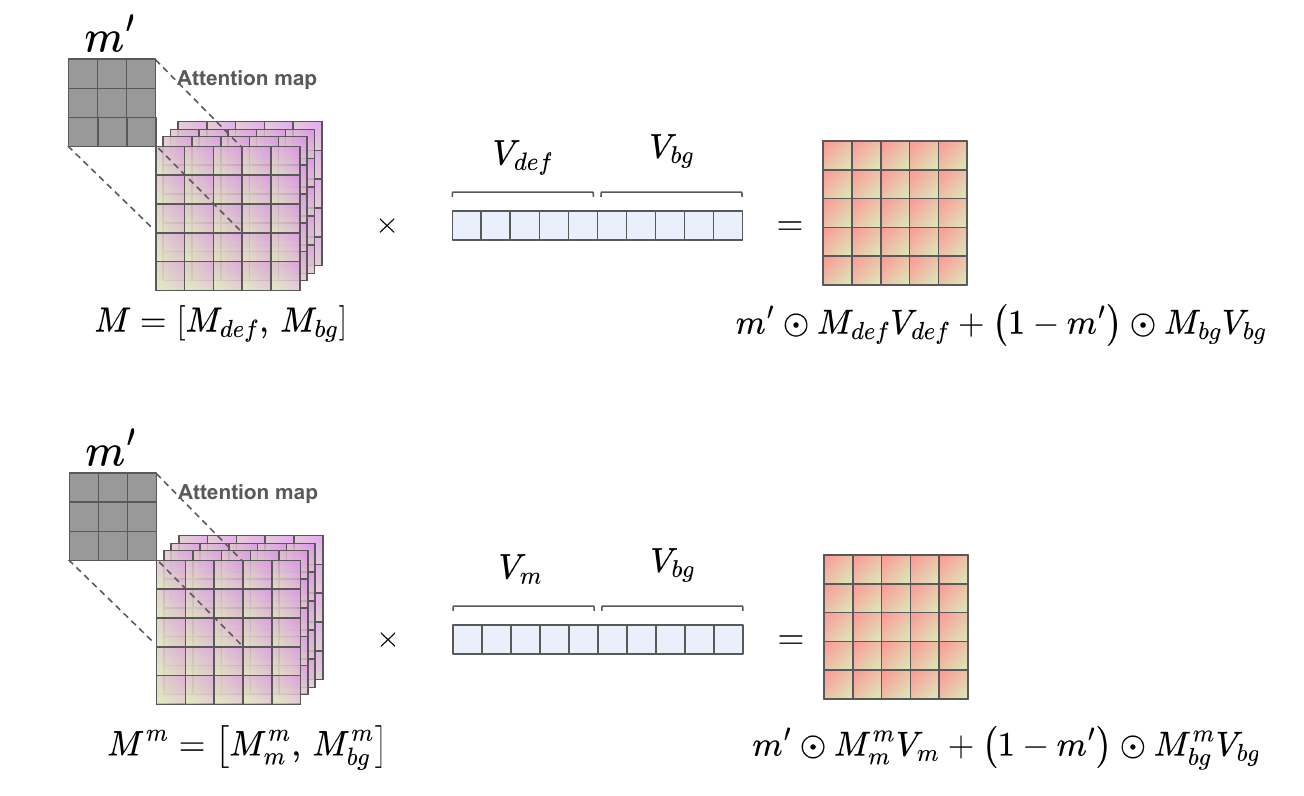}}
    \caption{Framework of our defect generation. The left image is the overview of the denoising process for latent in U-Net. The right image is the details of cross attention process in U-Net, where we use the masking strategy for disentangling each text embedding.}
    \label{fig:framework}

\end{figure*}

\subsection{Diffusion model}
Since the advent of diffusion models \citep{ho2020denoisingdiffusionprobabilisticmodels, song2021scorebased}, text-to-image (T2I) models \citep{Rombach_2022_CVPR} that control image generation through text has been developed. The denoising network $\epsilon_{\theta}$ is conditioned on text embedding $C$, where text embedding controls the output with cross-attention. Latent variable, $z_t$ is injected by Gaussian noise $\epsilon$ from the $z_0$ as follows \citep{ho2020denoisingdiffusionprobabilisticmodels}: 
\begin{equation}
    \label{eq:0}
    z_t=\sqrt{\alpha_t}z_0+\sqrt{1-\alpha_t}\epsilon, \epsilon \sim \mathcal{N}(0,I)
    \end{equation}

From noising the latent in Eq.\ref{eq:0}, the network is trained to minimize the following loss between random noise $\epsilon$ and the noise prediction $\epsilon_{\theta}(z_t,t, C)$: 
\begin{equation}
    \label{eq:1}
    \min_{\theta} E_{z_{0},\epsilon \sim \mathcal{N} (0,I), t \sim Uniform(1,T) } ||\epsilon - \epsilon_{\theta}(z_t,t,C)||_2^2     
\end{equation}

\subsection{DDIM Inversion}

By the DDIM Sampling \citep{song2021denoising}, we can denoise the noise with non-Markovian deterministic diffusion process as follows:
\begin{align}
    \label{eq:2}
    z_{t-1} &=\sqrt { \alpha_{t-1}}  ({ z_t -\sqrt{1-\alpha_t} \cdot \epsilon_{\theta}(z_t,t,C)  \over \sqrt{\alpha_t}}  ) 
 \notag \\ &+ \sqrt{1-\alpha_{t-1}  }\cdot\epsilon_{\theta}(z_t,t,C) \notag  \\ &=\sqrt{\alpha_{t-1} }   [ \sqrt{ 1\over \alpha_t}z_{t}+  (\sqrt{{ 1 \over \alpha_{t-1} }-1} - \sqrt{{1 \over \alpha_t} -1 }  )  \notag  \\ & \cdot \epsilon_{\theta}(z_t,t,C) ]
\end{align}

With approximating to ODE path, we can reverse the denoising steps (i.e., DDIM Inversion), where some initial $\tilde{z}_{T}$ can noised from the $z_{0}$ as follows: 
\begin{align}
    \label{eq:3}
    \tilde{z}_{t} &=\sqrt{\alpha_{t} }  [\sqrt{ 1 \over \alpha_{t-1}}\tilde{z}_{t-1}+ (\sqrt{{ 1 \over \alpha_{t} }-1} - \sqrt{{1 \over \alpha_{t-1}} -1 }  )  \notag \\ & \cdot\epsilon_{\theta}(\tilde{z}_{t-1},t-1,C)  ]
\end{align}

Especially, from the DDIM Inversion \citep {song2021denoising}, editing of target images is enabled with changing text embeddings \citep{hertz2023prompttoprompt, Mokady_2023_CVPR}. Furthermore, target images can be translated to source image concepts via controlling cross-attention. In this paper, we utilize the DDIM Inversion to personalize the target normal images and generate the defect in the target area.

\section{Method}

\subsection{Disentanglement Loss formulation}

% As previously mentioned, our goal is to model the relation between defect and background to generate more faithful defects on the normal image. In U-Net, attention interacts between the defect and background area to denoise the random noise. Although defects can vary depending on the background, the background should be independent of generating the defect. To disentangle the effect of the defect on the background, we propose loss formulation as follows. 

As discussed earlier, our objective is to model the relationship between defect and background in order to generate more accurate defect representations on normal images. In the U-Net architecture, attention mechanisms facilitate interaction between defect and background regions, effectively reducing random noise. While defects may vary depending on the background context, the background itself should remain unaffected by defect generation. To decouple the influence of defects on the background, we propose a disentanglement loss that ensures defect generation is conditioned on the background while preventing interference with background reconstruction.
\begin{align}    
     \label{eq:4}
     \mathcal{L}(\theta)    \notag  & \coloneq E_{z_{0},\epsilon, t } \underbrace{||m\odot(\epsilon - \epsilon_{\theta}(z_t,t,C_1))||_2^2}_{Matching \ Loss}  \\  \notag &+ \underbrace{||(1-m)\odot(\epsilon-\epsilon_{\theta}(z_t^{m},t,C_2))||_2^2}_{Regularizer} \\ &= ||\epsilon- m\odot\epsilon_{\theta}(z_t,t,C_1)-(1-m)\odot\epsilon_{\theta}(z_t^{m},t,C_2))||_2^2
\end{align}

We suppose that text embedding, $C_1=[C_{def}, C_{bg}]$, $C_2=[C_{m}, C_{bg}]$ and we mask its cross-attention with source mask, $m$ in the training dataset, where the $C_{def}$, $C_{m}$ attentions on masked area and $C_{bg}$ attentions on another area like Figure. \ref{fig:framework}. Masked latent is defined as masking the anomaly latent as follows: $z_0^m \coloneq (1-m) \odot z_0$, $z_0 \sim p(z_0)$. From Eq.\ref{eq:0}, we inject the noise to each latent as follows.
\begin{align}
    \label{eq:5}
    z_t &=\sqrt{\alpha_t}z_0+\sqrt{1-\alpha_t}\epsilon \\
    z_t^m  &=\sqrt{\alpha_t}z_0^m+\sqrt{1-\alpha_t} (1-m) \odot \epsilon
\end{align}
In Eq.\ref{eq:4}, minimizing $Matching \ Loss $ denoises that defect area of $z_t$ with considering the noised background area. Since $(1-m)\odot z_t = z_t^m$ is satisfied for all $t$, minimizing $Regularizer$ term denoises the background area of $z_t$, where the denoising process of the background area of $z_t$ is independent on the defect area of $z_t$. We prove this statement in Lemma \ref{lemma:1}.

\begin{lemma}{Suppose that \(\mathcal{L}(\theta^*)=0\), then background of $z_t$ is reconstructed as follows:
$(1-m)\odot z_{0}=(1-m)\odot \sqrt{1 \over \alpha_t}[z_t^m
-\sqrt{1-\alpha_t} \epsilon_{\theta^*}(z_t^m,t,C_2) ]$
\label{lemma:1}
}
\end{lemma}
According to Lemma \ref{lemma:1}, it is possible to reconstruct the background without requiring information about the defect area. As a result, the model can produce a variety of defects while generating the same background. The proof of Lemma \ref{lemma:1} is provided in Appendix \ref{proof:0}.

\subsection{DDIM Inversion for generating synthetic data}

To effectively control the defect on target normal latent, $z_{0}^{normal}$, it is essential to identify the initial state of the target normal latent representation. To achieve this, we employ DDIM Inversion on normal images using the target masks. This approach allows us to accurately initialize the latent space for normal images, enabling precise control over defect generation. From Eq.\ref{eq:3}, we reverse the denoising steps of $z_{0}^{m} :=(1-m) \odot z_{0}^{normal}$, $z_{0} \sim p(z_{0}^{normal})$ as follows:
\begin{align}
    \label{eq:6}
    \tilde{z}_{t}^{m}& =\sqrt{\alpha_{t} } \big[\sqrt{ 1 \over \alpha_{t-1}}\tilde{z}_{t-1}^{m}+(1-m) \odot (\sqrt{{ 1 \over \alpha_{t} }-1} \notag \\ &- \sqrt{{1 \over \alpha_{t-1}} -1 } )\epsilon_{\theta}(\tilde{z}_{t-1}^{m},t-1,C_2)  \big]
\end{align}
After reversing the latent, $\tilde{z}_T^{m}$, the denoising path should generate the defect on the mask area. To guarantee defect generation on the target mask, we present the following theorem to demonstrate that the standard initialization strategy works. We prove Theorem \ref{theorem:1} in Appendix \ref{proof:1}.    

% \begin{proposition}{Suppose that \(\mathcal{L}(\theta^*)=0\) and $\epsilon \sim N(0,I)$, then $\lim_{\alpha_t \rightarrow 0} z_t=z_t^m+m \odot \epsilon$ 
% }\label{prop:1}

% \end{proposition}
\begin{theorem}{Suppose that \(\mathcal{L}(\theta^*)=0\) and $\alpha_{t-1}>\alpha_t$ for all $t$ , then $\lim_{\alpha_t \rightarrow 0} \tilde{z_{t}}-\tilde{z_t}^m=m \odot \epsilon$, $\epsilon \sim \mathcal{N}(0,I)$
}\label{theorem:1}
\end{theorem}

Suppose that Theorem \ref{theorem:1}, we randomly initialized the latent, $z_T^*=\tilde{z}_T^{m}+m\odot\epsilon$, where $\epsilon \sim \mathcal{N}(0,I)$. From Eq.\ref{eq:7}, denoising $z_t^*$ generates the defect on the mask area affected by the denoised background part of latent, $z_t^m$.
\begin{align}
    \label{eq:7}
    z_{t-1}^{*} &= \sqrt{\alpha_{t-1}} [\sqrt{\frac{1}{\alpha_{t}}} z_{t}^{*} + (\sqrt{\frac{1}{\alpha_{t-1}} - 1} - \sqrt{\frac{1}{\alpha_{t}} - 1} ) \notag \\ 
    &  \times ( (1-m) \odot \epsilon_{\theta}(z_{t}^{m}, t, C_2) + m \odot \epsilon_{\theta}(z_{t}^{*}, t, C_1) ) ]
\end{align}
Although we initialize the masked area with random noise, Proposition \ref{prop:1} proves that denoising $z_t^*$ maintains the DDIM-Inversion gap, where we kept the background of the target normal image. 
\begin{proposition}{ Suppose that $\mathcal{L}(\theta^*)=0$ and DDIM Inversion Path follows Eq.\ref{eq:6} then $||(1-m)\odot(z_t^m-\tilde{z}_{t}^{m})||_2^2=||(1-m)\odot(z_t^*-\tilde{z}_{t}^{m})||_2^2$}
    \label{prop:1}
\end{proposition}
From Proposition \ref{prop:1}, defect generation on target mask still does not affect the denoising of background after DDIM Inversion. We prove Proposition \ref{prop:1} in Appendix \ref{proof:2}.

\subsection{Masking attention and refinement of mask }

\subsubsection{Masking attention strategy}

In the Text-to-Image (T2I) model, U-Net is designed with self-attention and cross-attention mechanisms, where cross-attention facilitates interaction between text embeddings and latent features. 

To disentangle each text embedding, we define masked cross-attentions of $z_t$ and $z_t^m$ inspired by \citep{park2024shape}. In the process of cross-attention, $m'$ is resized mask and each attention map is defined as follows: $M:=[M_{def}, M_{bg}]=[softmax({QK_{def}^T \over \tau} ), softmax({QK_{bg}^T \over \tau} )]$ and $M^m:=[M_{m}^m, M_{bg}^m]=[softmax({Q^mK_{m}^T \over \tau} ), softmax({Q^mK_{bg}^T \over \tau} )]$,
where $Q$ and $Q^m$ are from query vectors from $z_t$ and $z_t^m$, respectively. $K_{def}$, $K_{bg}$, and $K_{m}$ are the key vectors from the text encoder embeddings $C_{def}$, $C_{bg}$, and $C_{m}$, respectively.
Then, the masked cross-attention is given by
% $M^1:=[M_{def}^1, M_{bg}^1]=[softmax({Q^1K_{def}^T \over \tau} ), softmax({Q^1K_{bg}^T \over \tau} )]$ and $M^2:=[M_{m}^2, M_{bg}^2]=[softmax({Q^2K_{m}^T \over \tau} ), softmax({Q^2K_{bg}^T \over \tau} )]$.
% \begin{align}
%     \label{eq:8}
%     Attn^C(z_t,C_{1}) &=m' \odot Softmax({QK_{def}^T\over \tau })V_{def}
%     \notag \\  &+(1-m') \odot Softmax({QK_{bg}^T\over \tau })V_{bg} \\
%     Attn^C(z_t^m,C_{2}) &=m' \odot Softmax({QK_{m}^T\over \tau })V_{m} \notag \\  &+(1-m') \odot Softmax({QK_{bg}^T\over \tau })V_{bg}
% \end{align}
% \begin{align}
%     \label{eq:8}
%     Attn^C(z_t,C_{1}) &=m' \odot Softmax({QK_{def}^T\over \tau })V_{def}
%     \notag \\  &+(1-m') \odot Softmax({QK_{bg}^T\over \tau })V_{bg} \\
%     Attn^C(z_t^m,C_{2}) &=m' \odot Softmax({QK_{m}^T\over \tau })V_{m} \notag \\  &+(1-m') \odot Softmax({QK_{bg}^T\over \tau })V_{bg}
% \end{align}
% \begin{align}
%     \label{eq:8}
%     Attn^C(z_t,C_{1}) &=m' \odot Softmax({QK^T\over \tau })V_{def}
%     \notag \\  &+(1-m') \odot Softmax({QK^T\over \tau })V_{bg} \\
%     Attn^C(z_t^m,C_{2}) &=m' \odot Softmax({QK^T\over \tau })V_{m} \notag \\  &+(1-m') \odot Softmax({QK^T\over \tau })V_{bg}
% \end{align}
% \begin{align}
%     \label{eq:8}
%     Attn^C(z_t,C_{1}) &=m' \odot M_{def}^1 V_{def}+(1-m') \odot M_{bg}^1 V_{bg} \\
%     Attn^C(z_t^m,C_{2}) &=m' \odot M_{m}^2 V_{m}+(1-m') \odot M_{bg}^2 V_{bg}
% \end{align}
\begin{align}
    \label{eq:8}
    Attn^C(z_t,C_{1}) &=m' \odot M_{def} V_{def}+(1-m') \odot M_{bg} V_{bg} \\
    Attn^C(z_t^m,C_{2}) &=m' \odot M_{m}^m V_{m}+(1-m') \odot M_{bg}^m V_{bg}
\end{align}
with value vectors $V_{def}$, $V_{bg}$, and $V_{m}$ given like the key vectors.

Although we do not directly mask the self-attention layer, our formulation enables masked self-attention for the background part due to $m \odot z_t^m = 0 $ and $z_t^m =(1-m)\odot z_t^m $. Namely, the following masked self-attention holds:
\begin{align}
    \label{eq:9}
    Attn^S(z_t^m) \notag & =Softmax({\hat{Q}^m \hat{K}^{mT}\over \tau })\hat{V}^m \\ 
       \notag &=Softmax({\hat{Q}^m\hat{K}^{mT}\over \tau })(1-m') \odot \hat{V}^m \\ 
       &=(1-m')\odot Attn^S(z_t^m) 
\end{align}
Here, $\hat{Q}^m$, $\hat{K}^m$, $\hat{V}^m$ are the query, key, values on the self-attention layer from the masked latent $z_t^m$.
Since $(1-m) \odot \epsilon_{\theta} (z_t^m,t,C_2)$ denoises that the background of $z_t$, Eq.\ref{eq:9} shows that its denoising step only depends on the background of $z_t$.  
The details will be given in Appendix \ref{Appendix: attention}.

\subsubsection{Refining target mask with attention}
During inference, we guide defect generation within the target mask \( m \); however, its generation may not be filled within the target mask. This occurs because background latents can influence the defect region, occasionally incorporating background elements into the generated defect. To address this, we refine the target mask to ensure a more precise defect localization. This refined mask is then used to construct a robust training dataset for anomaly detection.

% In the inference step, we guide the generation of a defect on the target mask $m$, but its generation may not be fully contained within the target mask. Since the latent part of the background affects the defect area, generation sometimes contains the background part. Therefore, we refine the target mask for constructing a robust dataset, where it is used as a training dataset in anomaly detection. 

After DDIM Inversion to $\tilde{z}_T^*$ in Eq. \ref{eq:7}, we can save cross-attention map of $\epsilon_{\theta} (z_t^*,C_2)$, where text embedding of $C_{def}$ can refine the target mask $m$ with its attention map as Eq.\ref{eq:10}. Since the resolution of the cross-attention map is less than or equal to the resolution of latent, we should upscale the attention map to refine the target mask. $Resize$ means that the upscaling algorithm such as the Bilinear interpolation. 
\begin{equation}
\label{eq:10}
\begin{cases} 
    m^*_{i,j}=1, & \text{if } Resize \left(m' \odot Softmax({QK_{def}^T\over \tau })\right)_{i,j} >= {1 \over 2}  \\
    m^*_{i,j}=0, & \text{if }Resize \left(m' \odot Softmax({QK_{def}^T\over \tau })\right)_{i,j}< {1 \over 2}  
\end{cases}
\end{equation}

\begin{algorithm}[h!]
    \caption{Training and Inference}
    \label{alg:Training}
\begin{algorithmic}[1]
   \State {\bfseries Input}: Anomaly Dataset $\{z_{0},m\} \sim \{p(z_0), p(m)\}$, Normal Dataset $\{z_{0}^{normal},m\} \sim \{p(z_0^{normal}), p(m)\}$ U-Net parameter $\theta$, Context vector $C_1,C_2$ 
    \While{ not converged}  \Comment{Training}
    \State  Compute $\mathcal{L}(\theta)$ based on $\{z_0, m\}$
    \State Update $\theta \leftarrow \theta -h \nabla_{\theta} \mathcal{L}(\theta)$
    \EndWhile 
    \State DDIM Inversion for $z_0^m:=(1-m) \odot z_0^{normal}$ by Eq. \ref{eq:6} \Comment{Inference}
    \State DDIM Sampling for $z_T^*$ by Eq. \ref{eq:7}
    \State Mask Refinement for $m^*$ by Eq. \ref{eq:10}
    \State Return $z_0^*,m^*$
\end{algorithmic}
\end{algorithm}

Algorithm \ref{alg:Training} indicates our defect generation process, where synthetic data would be used as a training dataset for downstream anomaly detection tasks.

\section{Experiment}

\subsection{Experimental settings}

\textbf{Dataset\quad} In this paper, we conduct our experiments on industrial anomaly detection benchmark datasets: MVTec-AD \citep{bergmann2019mvtec} and MVTec-Loco \citep{bergmann2022beyond}. Each dataset includes anomaly instances in the test set, while the training set contains only normal instances. To train the generative model on these anomaly instances, we split each anomaly dataset into two folds: one fold is used for training the generative model, and the other for evaluating inspection performance in anomaly detection. For unsupervised settings, we exclude anomaly instances from training, as unsupervised methods do not learn from anomaly data.

% In this paper, our experiments are conducted on industrial anomaly detection benchmark datasets; MVTec-AD \citep{bergmann2019mvtec} and MVTec-Loco \citep{bergmann2022beyond}. Each dataset contains the anomaly instances in test dataset, whereas the train set contained only normal instances. To learn the generative model with these anomaly instances, we split each anomaly dataset into two-folds; one fold to train the generative model and the other one to test inspection performances for anomaly detection. Since unsupervised-based methods do not learn the anomaly instances in training, we exclude the anomaly instances for unsupervised settings. 

\noindent \textbf{Details of experiments\quad} We utilize the Stable Diffusion \citep{rombach2022high} for fine-tuning anomaly dataset with 500 iterations, where each model is trained on the defect type of objects. We only trained the U-Net part of Stable Diffusion followed by Dreambooth \citep{Ruiz_2023_CVPR} and inferred the sample with DDIM sampler \citep{song2021denoising} with 50-time steps.

\noindent \textbf{Baseline\quad} We compared our methodology with other defect generation-based methodologies; DFMGAN \citep{duan2023few} and Anomalydiffusion \citep{hu2024anomalydiffusion}, where each setting is identical in their papers. In addition, we compared ours with Unsupervised-based anomaly detection; RD4AD \citep{deng2022anomaly}, Patchcore \citep{roth2022towards}, and SimpleNet \cite{liu2023simplenet}. 

\subsection{Synthetic Data Quality}

For comparing generation quality between ours and other baselines, we measure FID score \citep{heusel2017gans} and LPIPS \citep{zhang2018unreasonable} for each category in MVTec-AD and MVTec-Loco, and use 100 synthetic images per each categories from each generative model trained on split folds in each benchmark datasets. Although previous benchmarks do not consider
Table \ref{tab:quality_ad} demonstrates our methodology can generate more realistic and diverse structure anomaly images, where our method overall outperforms the baseline. In addition, Table \ref{tab:quality_loco} shows that our methodology outperforms in logical anomalies, where our model generates the defect considering the background. 

Since DFMGAN does not control the generation given the target mask, its generation is not as diverse as diffusion-based methods and its generation quality is low overall. 
\begin{table}[h!]
\caption{Result of FID Score and LPIPS in MVTec-AD. Bold means the best performance and underline means the second-best performance. }
    
    \centering
     \resizebox{\columnwidth}{!}{
        \begin{tabular}{c|cc|cc|cc|} 
            \toprule   
            \multirow{2}{*}{\textbf{Category}}&\multicolumn{2}{c}{DFMGAN}&\multicolumn{2}{c}{AnomalyDiffusion}&\multicolumn{2}{c|}{Ours}\\ \cmidrule(lr){2-7}
             &FID$\downarrow$ &LPIPS$\downarrow$ & FID$\downarrow$ &LPIPS$\downarrow$ & FID$\downarrow$ &LPIPS$\downarrow$  \\ \midrule
            \multirow{1}{*}{Bottle} & 77.23 &0.176  & \underline{57.09}& \underline{0.150}  &  \textbf{52.23} &\textbf{0.143}  \\ 
            \multirow{1}{*}{Cable} &  199.22 & 0.461  & \underline{136.00}&\underline{0.402}&  \textbf{125.40} & \textbf{0.396}\\ 
            \multirow{1}{*}{Capsule} & 58.19 &0.191    &40.63 &\textbf{0.142} &  \textbf{28.21} & \underline{0.143} \\ 
            \multirow{1}{*}{Carpet}  & \textbf{44.56}& \textbf{0.273}     &64.29&\textbf{0.273} &53.92 & 0.277 \\ 
            \multirow{1}{*}{Grid} & 186.63 &\underline{0.444}  &\underline{135.10} &0.451&   \textbf{131.38}& \textbf{0.433}\\ 
            \multirow{1}{*}{Hazelnut} & 89.40 &0.305    &\underline{82.37}&\underline{0.260} &  \textbf{69.20} & \textbf{0.253} \\ 
            \multirow{1}{*}{Leather}  & \textbf{130.28}& 0.372 & 161.29&\textbf{0.346} &  \underline{141.34} & \underline{0.360} \\ 
            \multirow{1}{*}{Metal Nut} & \underline{85.89}  & 0.357    & 97.50&\underline{0.332}&  \textbf{76.58} & \textbf{0.315}\\ 
            \multirow{1}{*}{Pill}  & 177.87& 0.280  & \underline{67.73}&\underline{0.202}&  \textbf{56.85} & \textbf{0.194}\\ 
            \multirow{1}{*}{Screw} & \textbf{34.46} & \underline{0.353}   &39.61 &\textbf{0.338} &  \underline{36.52} &  \textbf{0.338}\\ 
            \multirow{1}{*}{Tile}  & 232.08&0.519    &\underline{205.33}&\textbf{0.509}&  \textbf{134.77} & \underline{ 0.510}\\ 
            \multirow{1}{*}{Toothbrush} &  98.10&0.246   & \underline{57.63}&\underline{0.159} &  \textbf{49.94} &\textbf{ 0.158}\\ 
            \multirow{1}{*}{Transistor} & 132.70 &0.377   &\underline{99.36} &\underline{0.338}&  \textbf{92.41} & \textbf{0.328}\\ 
            \multirow{1}{*}{Wood}  & 211.87& 0.398    &\underline{158.58}&\underline{0.357} &   \textbf{121.45}&\textbf{0.353} \\ 
            \multirow{1}{*}{Zipper}  & 116.46 & 0.264    & \underline{98.02}& \underline{0.255} &  \textbf{ 91.09} & \textbf{0.249} \\ 
            \bottomrule 
            \multirow{1}{*}{Average}&132.96  &0.334 &\underline{104.36} &\underline{0.301} &  \textbf{88.17} & \textbf{0.297} \\
            \bottomrule
        \end{tabular}
}
\label{tab:quality_ad}
\end{table}

\begin{table*}[t!]
    \centering

         \caption{Result of anomaly detection in benchmark datasets. We ensure a fair comparison by keeping the dataset configuration of the unsupervised baselines the same as for the synthetic-based baselines, except for the training dataset. Notably, the synthetic-based baselines use supervised settings, where their training datasets include anomaly instances. For MVTec-Loco, we only conduct experiments for logical anomaly class. 
         Bold text indicates the best performance, and underlined text highlights the second-best performance.}
        \begin{minipage}{0.675\textwidth}
            \centering
           \resizebox{\linewidth}{!}{\begin{tabular}{c|ccc|cccc} 
            \toprule
            \multicolumn{8}{c}{Unsupervised-based} \\ 
                   \midrule 
        \multirow{2}{*}{Method} & \multicolumn{3}{c|}{Image-level} &\multicolumn{4}{c}{Pixel-level} \\                    
          & AUROC&AP&F1&AUROC&AP&F1&AUPRO \\ \midrule
          RD4AD&98.0 &99.0&97.8&97.8&56.7&59.9&\textbf{93.9} \\
        Patchcore&\underline{99.1} &\textbf{99.8}&\textbf{98.6}&\textbf{98.1}&57.3&59.9&93.0 \\
        SimpleNet&\textbf{99.3} &\underline{99.7}&\textbf{98.6}&97.7&55.1&57.4&91.4 \\

        \toprule
        \multicolumn{8}{c}{Synthetic-based} \\          \midrule 

      DFMGAN& 98.0&99.4&97.6&96.2&74.0&71.1&88.6 \\
        Anomalydiffusion& 98.0&99.3&\underline{97.9}&\underline{97.9}&\underline{77.1}&\underline{73.4}&91.7 \\ \midrule
        Ours &\textbf{99.3}&\textbf{99.8}&\textbf{98.6}&\textbf{98.1}&\textbf{78.9}&\textbf{74.2}&\underline{93.6} 
            \\
        \bottomrule
        \end{tabular} 
            }        
        \label{tab:localization_mvtec}
         \subcaption{MVTec-AD}

        \end{minipage}%

        \begin{minipage}{0.675\textwidth}
            \centering
             \resizebox{\linewidth}{!}{      \begin{tabular}{c|ccc|cccc}
            \toprule
            \multicolumn{8}{c}{Unsupervised-based} \\ 
                   \midrule 
        \multirow{2}{*}{Method} & \multicolumn{3}{c|}{Image-level} &\multicolumn{4}{c}{Pixel-level} \\                    
          & AUROC&AP&F1&AUROC&AP&F1&AUPRO \\ \midrule
        RD4AD&70.2 &63.6&61.8&72.4&21.6&25.4&57.6 \\
        Patchcore&69.1&62.4&60.4&70.6&33.3&33.4&55.2 \\
        SimpleNet& 74.0&67.3&62.8&71.5&35.8&36.8&53.4 \\
    
        \toprule
        \multicolumn{8}{c}{Synthetic-based} \\          \midrule 
    
        DFMGAN& 84.7&79.8&77.0&88.6&51.8&55.1&78.3 \\
        Anomalydiffusion& \underline{85.8}&\underline{81.9}&\underline{78.5}&\underline{91.4}&\underline{59.0}&\underline{61.2}&\underline{80.2} \\             \bottomrule
    
        Ours &\textbf{89.9}&\textbf{87.1}&\textbf{84.0}&\textbf{95.3}&\textbf{79.8}&\textbf{78.2}&\textbf{84.2}  \\   \bottomrule
        \end{tabular} }
        \label{tab:localization_mvtec_loco}
        \subcaption{MVTec-Loco}

    \end{minipage}

\end{table*}

\begin{figure}[h]
    \centering    
    \begin{subfigure}[h!]{\linewidth}
    \centering
    
    \includegraphics[width=0.48\linewidth]{figure/juice_ad.pdf}
    \includegraphics[width=0.48\linewidth]{figure/juice_ours.pdf}
    \includegraphics[width=0.48\linewidth]{figure/transistor_ad.pdf}
    \includegraphics[width=0.48\linewidth]{figure/transistor_us.pdf}
    
    \end{subfigure}
    \caption{Visualization of synthetic instances given identical target mask. The left side is from Anomalydiffusion and the right side is from ours.  The right side mask is refined mask from Eq. \ref{eq:10}.}
    \label{fig:Generation} 
    
\end{figure}
Although Anomalydiffusion \citep{hu2024anomalydiffusion} is also diffusion-based method, it sometimes generates some unrealistic defects depending on the target mask mentioned previously. Figure \ref{fig:Generation} shows that Anomalydiffusion generates the defects given target mask, but its generation can be located out of objects. Since they usually generate fully filled defects in the masked area, their generation could be low-quality, where its generation can be bias for anomaly detection training. 
Further generations are shown in Appendix \ref{Appendix:additional defect}. 
\begin{table}[h!]
\caption{Result of FID Score and LPIPS in MVTec-Loco. Bold means the best performance and underline means the second-best performance. }
    \centering
    \resizebox{\columnwidth}{!}{%

        \begin{tabular}{c|cc|cc|cc|} \toprule    
            \multirow{2}{*}{\textbf{Category}}&\multicolumn{2}{c}{DFMGAN}&\multicolumn{2}{c}{AnomalyDiffusion}&\multicolumn{2}{c|}{Ours}\\ \cmidrule(lr){2-7}
            &FID$\downarrow$ &LPIPS$\downarrow$ & FID$\downarrow$ &LPIPS$\downarrow$ & FID$\downarrow$ &LPIPS$\downarrow$ \\ \midrule
            \multirow{1}{*}{Breakfast Box}&178.61 & 0.419  & \underline{119.42}& \underline{0.353}  &  $\mathbf{84.90}$&$\mathbf{0.316}$\\ 
            \multirow{1}{*}{Screw Bag}&245.83  & 0.523   & \underline{116.19}&\underline{0.368}&  $\mathbf{104.62}$& $\mathbf{0.367}$\\ 
            \multirow{1}{*}{Juice Bottle} &\underline{71.49}  &  0.344  &76.17 & \underline{0.280} &  $\mathbf{44.94}$ & $\mathbf{0.246}$ \\ 
            \multirow{1}{*}{Splicing Connectors}& \textbf{90.87}  &  \textbf{0.402}  &125.71&0.497 & $\underline{114.61}$ & $\underline{0.483}$ \\ 
            \multirow{1}{*}{Pushpins}& 162.20  &  0.471   &\underline{56.05}&\underline{0.385}&   $\mathbf{44.53}$& $\mathbf{0.382}$\\ 
            \bottomrule 
            \multirow{1}{*}{Average}  & 145.19   & 0.432  &\underline{102.92} &\underline{0.376} &  $\mathbf{79.49}$ & $\mathbf{0.359}$ \\
            \bottomrule
            
        \end{tabular}
                        }    
\label{tab:quality_loco}

\end{table}
\begin{figure*}[h!]
    \centering    
     \resizebox{\linewidth}{!}{%
    \begin{subfigure}[h!]{0.3\linewidth}
    \centering
    \includegraphics[width=\linewidth]{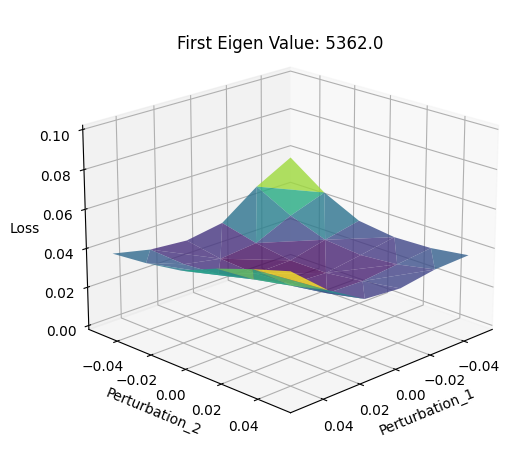}\\
    \includegraphics[width=\linewidth]{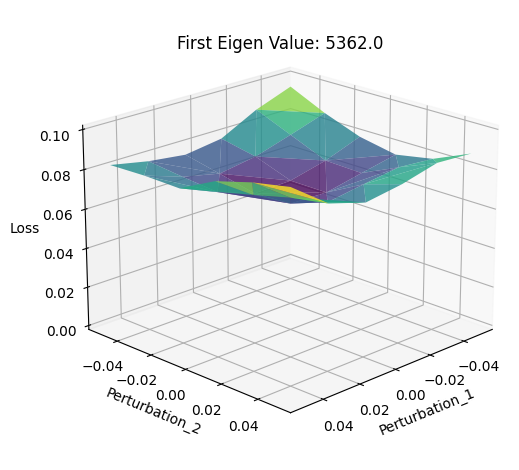}
    \caption{DFMGAN}
    \end{subfigure}
    \begin{subfigure}[h!]{0.3\linewidth}
    \centering
    \includegraphics[width=\linewidth]{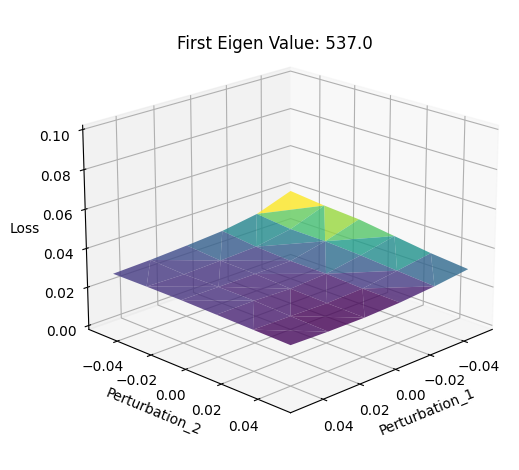}
\\
    \includegraphics[width=\linewidth]{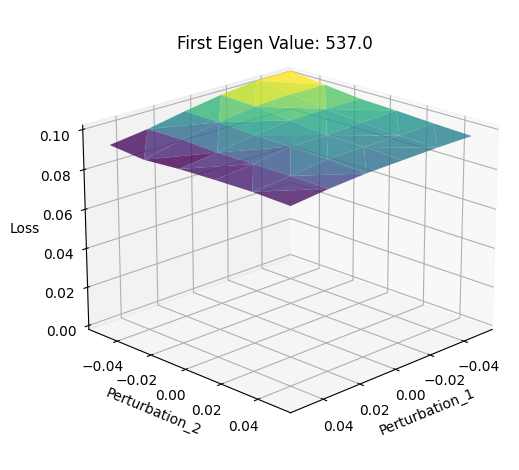}
    \caption{Anomalydiffusion}
    \end{subfigure}
    \begin{subfigure}[h!]{0.3\linewidth}
    \centering
    \includegraphics[width=\linewidth]{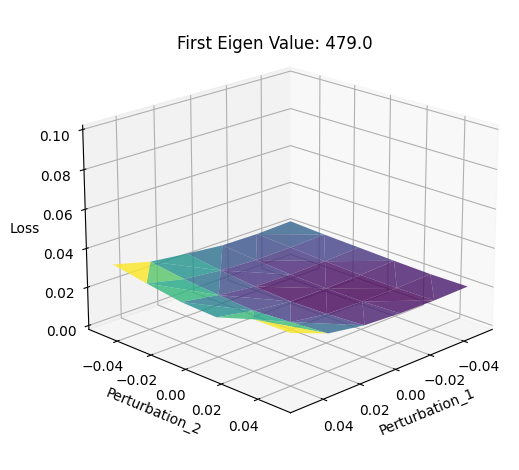}
\\
    \includegraphics[width=\linewidth]{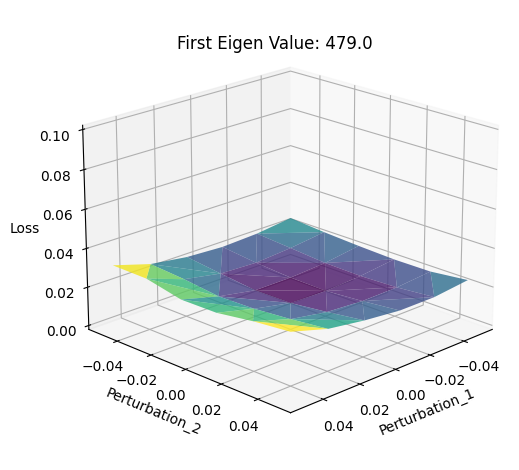}
    \caption{Ours}
    \end{subfigure}}
    \caption{Comparison of training loss landscape between ours and baselines in MVTec-AD. The first row is the loss landscapes of training normal sample. The second row is the loss landscapes of training anomaly sample except for synthetic anomalies.}
    \label{fig:loss_landscape} 
    
\end{figure*}
\subsection{Anomaly Detection}

\subsubsection{Quantitative Analysis}

To demonstrate our synthetic methodology's effectiveness in anomaly detection, we utilized the generated dataset for training a segmentation backbone. In this experiment, we trained naive U-Net \citep{ronneberger2015u} with both real and synthetic defect datasets. We use normal samples in training and 100 defect instances per defect type for 200 epochs followed by the framework of Anomalydiffusion \citep{hu2024anomalydiffusion}.
\\ 
For a fair comparison, we report the average performance of each split fold, where a test set of one fold is used as a train set of another fold. Furthermore, we also compare our method with the Unsupervised-based anomaly detection (UAD) baselines. To fairly report the performances of UAD, we limit the normal train set to be same with synthetic-based methods and report detection performances of each fold. The other settings of UAD baselines are identical to their origin paper.

Table \ref{tab:localization_mvtec} demonstrates that our synthetic dataset enhances both detection and segmentation performances of anomaly detection in MVTec-AD, where we report the AUROC, AP (Average Precision), and F1 max score for image and pixel level. Additionally, we report the PRO score to consider the inspection performance of tiny defects. Although image-level metrics of unsupervised-based methods attain higher scores than other synthetic-based methods, our method is comparable and even better in pixel-level performances.  
Furthermore, we conducted anomaly detection in the MVTec-Loco, where we reported the performance of logical anomalies. Table \ref{tab:localization_mvtec_loco} shows that our synthetic strategy outperforms other baselines in both image and pixel levels.  
Since unsupervised-based anomaly detection exploits the normal features, the detection performance shows degradation compared to the supervised setting.   
Although our methodology is based on the diffusion model, the performance demonstrates that our generation facilitates more robust training by reducing logically incorrect generations, as illustrated in Figure \ref{fig:Generation}, where it demonstrates that the quality of generation is related to the performance of anomaly detection. 
% \begin{table}[h!]
% \caption{Result of anomaly detection in MVTec-Loco. Bold means the best performance and underline means the second-best performance. }
%         \centering
%     \resizebox{\linewidth}{!}{
%          \begin{tabular}{c|ccc|cccc}
%             \toprule
%             \multicolumn{8}{c}{Unsupervised-based} \\ 
%                    \midrule 
%         \multirow{2}{*}{Method} & \multicolumn{3}{c|}{Image-level} &\multicolumn{4}{c}{Pixel-level} \\                    
%           & AUROC&AP&F1&AUROC&AP&F1&AUPRO \\ \midrule
%         RD4AD&70.2 &63.6&61.8&72.4&21.6&25.4&57.6 \\
%         Patchcore&69.1&62.4&60.4&70.6&33.3&33.4&55.2 \\
%         SimpleNet& 74.0&67.3&62.8&71.5&35.8&36.8&53.4 \\

%         \toprule
%         \multicolumn{8}{c}{Synthetic-based} \\          \midrule 

%         DFMGAN& 84.7&79.8&77.0&88.6&51.8&55.1&78.3 \\
%         Anomalydiffusion& 85.8&81.9&78.5&91.4&59.0&61.2&80.2 \\             \bottomrule

%         Ours &\textbf{89.9}&\textbf{87.1}&\textbf{84.0}&\textbf{95.3}&\textbf{79.8}&\textbf{78.2}&\textbf{84.2}  \\   \bottomrule
%         \end{tabular} 
%         }

%         \label{tab:localization_mvtec_loco}
% \end{table}

\subsubsection{Qualitative Analysis}

To illustrate how unrealistic synthetic data can compromise the robustness of anomaly detection, we extend our experiments by analyzing the impact of synthetic datasets through the lens of the training loss landscape \citep{li2018visualizing}. Specifically, we visualize the loss landscapes of normal and anomaly training datasets without synthetic instances in Figure. \ref{fig:loss_landscape}. Prior studies \citep{jiang2019fantastic, kim2023saal} suggest that achieving a flat minimum in the loss function is strongly associated with better model generalization. Following this insight, our approach facilitates more robust learning for detecting unseen anomalies. Since a lower first eigenvalue of the loss Hessian corresponds to a flatter loss landscape, our detection model exhibits superior generalization compared to other baselines, validating its effectiveness in capturing anomalies at both the image and pixel levels.

\subsubsection{Ablation Study}
\textbf{Effect of Regularizer in Eq. \ref{eq:4} \quad} To demonstrate our loss formulation effective in generating a faithful synthetic dataset, we compared our loss function with naive loss $L'$ defined as Eq.\ref{eq:1}. After minimizing $L'$, we denoise the random noise $z_T \sim \mathcal{N}(0,I)$ by DDIM sampler. Table \ref{tab:ablation_loss} shows that modeling the relation between background and defect can facilitate diversity and fidelity of generation. In addition, initializing the latent via DDIM inversion is more appropriate than random initialization in the few-shot training scenario.  

\begin{table}[h!]
\caption{Ablation study for loss function. We compare our loss formulation with naive matching loss and report the sampling quality of the synthetic dataset in MVTec-AD.}
    \centering
     \resizebox{0.8\columnwidth}{!}{
        \begin{tabular}{c|cc|cc|} 
            \toprule   
            \multirow{2}{*}{\textbf{Category}}&\multicolumn{2}{c}{$L'(\theta)$}&\multicolumn{2}{c|}{$L(\theta)$}\\ \cmidrule(lr){2-5}
             &FID$\downarrow$ &LPIPS$\downarrow$ & FID$\downarrow$ &LPIPS$\downarrow$  \\ \midrule
            \multirow{1}{*}{Capsule}  &117.25 &0.294 &  \textbf{28.21} &\textbf{0.143} \\
            \multirow{1}{*}{Hazelnut}   &134.27&0.357 &  \textbf{69.20} & \textbf{0.253} \\ 
            \multirow{1}{*}{Transistor}   &162.84 &\textbf{0.312}&  \textbf{92.41} & 0.328\\ 
            \multirow{1}{*}{Wood}  &196.76&0.369 &   \textbf{121.45}&\textbf{0.353} \\ 

            \bottomrule 
            \multirow{1}{*}{Average}&152.78  &0.333 &  $\mathbf{77.82}$ & $\mathbf{0.269}$ \\
            \bottomrule
        \end{tabular}
}\label{tab:ablation_loss}
\end{table}
\textbf{Effect of Mask Refinement in Eq. \ref{eq:10} \quad} As shown in Figure \ref{fig:Generation}, our defect generation is affected by background information given an unseen mask and refines the target mask with cross-attention in Eq. \ref{eq:10}. To demonstrate the mask refinement trick, we conduct an ablation study in anomaly detection to check the performance gain. Table \ref{tab:ablation_mask} shows that the refinement strategy can alleviate noise label issues of synthetic data, which makes the detection model more robust. 

\begin{table}[h!]
    \caption{Ablation study for mask refinement. We report averaged detection performances of capsule, transistor, hazelnut, and a wood class of MVTec-AD.}

    \centering
         \resizebox{\columnwidth}{!}{

    \begin{tabular}{c|ccc|cccc}
    \toprule
         \textbf{Refinement}& \multicolumn{3}{c}{Image-level}& \multicolumn{4}{c}{Pixel-level} \\          & AUROC& AP& F1 &AUROC& AP&F1&AUPRO\\        \cmidrule(lr){1-8}     
          $\times$ & 98.4 &99.6  &98.4 &96.4&\textbf{80.6}&\textbf{77.4}&91.7\\                  
          \midrule
         $\circ$ &\textbf{99.8} &\textbf{100.0} &\textbf{99.7}&\textbf{98.3}&80.0&75.4&\textbf{94.1} \\ \bottomrule
    \end{tabular}
}
    \label{tab:ablation_mask}
\end{table}

\section{Conclusion}

% We propose a diffusion-based defect generation methodology to mitigate data imbalance in anomaly detection. To enhance controllability and quality of generation, we propose our masked noising strategy and disentanglement loss to make the background part independent of defect part. Furthermore, we theoretically show the independence of denoising background and the relationships between latent and masked latent. 

% Based on the theoretical statement, we demonstrate our methodology outperforms the generation quality in benchmark datasets. Furthermore, our modeling can mitigate unwanted defect generation, which differs greatly from real anomaly instances. 

% From the quality of generation, our synthetic instances enhance anomaly detection performances compared with unsupervised-based anomaly detection. In addition, we demonstrate our synthetic instances make the loss landscapes flatter, where the detection model can be generalized. 

% In addition, we conduct ablation studies for demonstrating our regularizer and refinement of mask can enhance quality of generation and robustness of detection model. 

We propose a diffusion-based defect generation methodology to address data imbalance in anomaly detection. To improve both the controllability and quality of defect generation, we introduce a masked noising strategy along with a disentanglement loss, which ensures that the background is independent of the defect. Furthermore, we provide theoretical evidence supporting the independence of background denoising and the relationships between the defect latent and masked latent representations.

Based on this theoretical foundation, we show that our methodology outperforms existing approaches in terms of generation quality on benchmark datasets. Our model also helps mitigate the generation of unrealistic defects, which significantly differ from real anomaly instances.

From a generation quality perspective, our synthetic anomaly instances improve anomaly detection performance compared to unsupervised-based methods. Additionally, we demonstrate that these synthetic instances help flatten the loss landscapes, enabling better generalization of the detection model.

Finally, we conduct ablation studies to validate that our regularizer and mask refinement strategies enhance both the quality of defect generation and the robustness of the detection model.

\newpage

%성능 이야기, ablation이야기, 등등 적기. 

\bibliography{example_paper}
\bibliographystyle{icml2025}

%%%%%%%%%%%%%%%%%%%%%%%%%%%%%%%%%%%%%%%%%%%%%%%%%%%%%%%%%%%%%%%%%%%%%%%%%%%%%%%
%%%%%%%%%%%%%%%%%%%%%%%%%%%%%%%%%%%%%%%%%%%%%%%%%%%%%%%%%%%%%%%%%%%%%%%%%%%%%%%
% APPENDIX
%%%%%%%%%%%%%%%%%%%%%%%%%%%%%%%%%%%%%%%%%%%%%%%%%%%%%%%%%%%%%%%%%%%%%%%%%%%%%%%
%%%%%%%%%%%%%%%%%%%%%%%%%%%%%%%%%%%%%%%%%%%%%%%%%%%%%%%%%%%%%%%%%%%%%%%%%%%%%%%
\newpage
\appendix
\onecolumn
\section{Appendix}

\subsection{Details of Training}

\textbf{Details of datasets} In our experiments on data generation and anomaly detection, we use the MVTec-Ad and MVTec-Loco datasets. Although the training sets for both datasets consist exclusively of normal samples, we establish a supervised setting by splitting the anomaly test set into two folds. Specifically, one fold contains the first half of the anomaly test samples, while the other fold contains the second half. 
During training, we first train the diffusion model using one fold of anomalies and perform inference using the normal samples from the training set. Next, we train the U-Net using the original normal training samples, the anomalies from one fold, and synthetic anomalies. 
For evaluation, we assess the model's performance on the remaining fold of anomalies and the normal test samples. For the unsupervised-based methods, we just utilize normal samples for training, which are identical settings with their paper.

\noindent \textbf{Details of training diffusion model} In our framework, we utilize the stable diffusion with $256 \times 256$ resolutions. We train the only U-Net part with 500 iterations per each class of objects.   

\noindent \textbf{Details of Inference} After training the diffusion model with one-fold anomalies, we utilize the DDIM inversion with 50 time steps to train normal samples and generate anomalies with a ground-truth mask in one-fold.

\subsection{Proof of Lemma \ref{lemma:1}}

\begin{lemma}{Suppose that \(\mathcal{L}(\theta^*)=0\), then background of $z_t$ is reconstructed as follows:
\\ 
\((1-m)\odot z_{0}= (1-m)\odot \sqrt{1 \over \alpha_t}\left[z_t ^m
-\sqrt{1-\alpha_t} \epsilon_{\theta^*}( z_t^m,t,C_2)\right]\)}

\end{lemma}

\begin{proof}
    \label{proof:0}
    \begin{align}
        \mathcal{L}(\theta^*)&=||\epsilon- m\odot\epsilon_{\theta^*}(z_t,t,C_1)-(1-m)\odot\epsilon_{\theta^*}(z_t^{m},t,C_2))||_2^2
       \notag \\ &=0  \\ 
        \epsilon &= m\odot\epsilon_{\theta^*}(z_t,t,C_1)+(1-m)\odot\epsilon_{\theta^*}(z_t^{m},t,C_2))     
    \end{align}
   From Eq.\ref{eq:5}, we suppose that the optimal denoised equation of $z_t$ as follows. 
    \begin{align}
        z_0&=  \sqrt{1 \over \alpha_t} \left [ z_t -\sqrt{1-\alpha_t}  \epsilon \right] \\
        &= \sqrt{1 \over \alpha_t} \left [ z_t -\sqrt{1-\alpha_t} ( m\odot\epsilon_{\theta^*}(z_t,t,C_1)+(1-m)\odot\epsilon_{\theta^*}(z_t^{m},t,C_2)) \right]  
    \end{align}
   Due to  $z_t^m = (1-m) \odot z_t  \ \forall t $, we prove this statement. 
    \begin{align}
        (1-m) \odot z_0&=  (1-m) \odot \sqrt{1 \over \alpha_t} \left [ z_t -\sqrt{1-\alpha_t}  \epsilon \right] \\
        &= \sqrt{1 \over \alpha_t} \left [ (1-m) \odot z_t -\sqrt{1-\alpha_t} ((1-m)\odot\epsilon_{\theta^*}(z_t^{m},t,C_2)) \right] \\
        &= \sqrt{1 \over \alpha_t} \left [ z_t^m -\sqrt{1-\alpha_t} ((1-m)\odot\epsilon_{\theta^*}(z_t^{m},t,C_2)) \right] 
    \end{align}

    % $\epsilon= m\odot\epsilon_{\theta^*}(z_t,t,C_1)+(1-m)\odot\epsilon_{\theta^*}(z_t^{m},t,C_2))$ and $z_0={1\over \sqrt{\alpha_t}}\left[z_t -\sqrt{1-\alpha_t} \epsilon \right]$. Since $(1-m) \odot z_t =z_t^m$ is satiesfied for all $t$, \((1-m)\odot z_{0}={1\over \sqrt{\alpha_t}}\left[(1-m)\odot z_t-\sqrt{1-\alpha_t} \epsilon_{\theta^*}((1-m)\odot z_t,t,C_2)\right]\) 
   
\end{proof}

\subsection{Proof of Theorem \ref{theorem:1}}

\begin{theorem}{Suppose that \(\mathcal{L}(\theta^*)=0\) and $\alpha_t<\alpha_{t-1}$ for all $t$ , then $\lim_{\alpha_t \rightarrow 0} \tilde{z_{t}}-\tilde{z_t}^m=m \odot \epsilon$, $\epsilon \sim \mathcal{N}(0,I)$}

\end{theorem}

\begin{proof}
    \label{proof:1}

%좀 더 수식을 구체화해서 적을 수 있지 않을까? 

% Since $\theta^*$ is also optimal solution of DDIM formulation from Theorem 1 of \citep{song2021denoising}, we can derive as follows:\\
Suppose that \(\mathcal{L}(\theta^*)=0\) and we reformulate DDIM sampling as Euler method followed by \citep{song2021denoising}.

\begin{equation}
    { z_{t-\Delta t} \over \sqrt{\alpha_{t-\Delta t }}} ={z_t \over \sqrt{\alpha_t} }  + \left ( \sqrt{ {1-\alpha_{t-\Delta t } \over \alpha_{t-\Delta t }}} -\sqrt{ {1-\alpha_{t} \over \alpha_{t}}}  \right ) \epsilon_{\theta^*} (z_t,t,C)
\end{equation}

For $\Delta t \rightarrow 0 $, we inverse the DDIM Sampling as follows: 

\begin{equation}
    z_t  = \sqrt{\alpha_{t} } [\sqrt{ 1 \over \alpha_{t-1}}z_{t-1}+ (\sqrt{{ 1 \over \alpha_{t} }-1} - \sqrt{{1 \over \alpha_{t-1}} -1 } )\cdot \epsilon_{\theta^*}(z_{t-1},t-1,C) 
\end{equation}

$\tilde{z}_t$ is forwarded by DDIM Inversion as follows:
\begin{equation}
    \tilde{z}_{t}  = \sqrt{\alpha_{t} } [\sqrt{ 1 \over \alpha_{t-1}}\tilde{z}_{t-1}+ (\sqrt{{ 1 \over \alpha_{t} }-1} - \sqrt{{1 \over \alpha_{t-1}} -1 } )\cdot((1-m) \odot \epsilon_{\theta^*}(\tilde{z}_{t-1}^{m},t-1,C_2) + m \odot \epsilon_{\theta^*}(\tilde{z}_{t-1}, t-1, C_1))]    
\end{equation}
 
 From \textbf{Theorem.1} of \citep{song2021denoising}, $m \odot \epsilon_{\theta^*}(\tilde{z}_{t-1}, t-1, C_1) = m \odot \epsilon $ is satisfied, then the following equations are derived.

\begin{align}
    \tilde{z}_{t}  &= \sqrt{\alpha_{t} } [\sqrt{ 1 \over \alpha_{t-1}}\tilde{z}_{t-1}+ (\sqrt{{ 1 \over \alpha_{t} }-1} - \sqrt{{1 \over \alpha_{t-1}} -1 } )\cdot((1-m) \odot \epsilon_{\theta}(\tilde{z}_{t-1}^{m},t-1,C_2) + m \odot \epsilon)] \\     
    \tilde{z}_{t}-\tilde{z}_{t}^m &=\sqrt{\alpha_{t} } [\sqrt{ 1 \over \alpha_{t-1}} m \odot \tilde{z}_{t-1}  + (\sqrt{{ 1 \over \alpha_{t} }-1} - \sqrt{{1 \over \alpha_{t-1}} -1 } )\cdot( m \odot \epsilon)]    
\end{align}

Therefore, $\lim_{\alpha_t \rightarrow 0} \tilde{z_{t}}-\tilde{z_t}^m=m \odot \epsilon$

\end{proof}

\subsection{Proof of Proposition \ref{prop:1}}
% \begin{theorem}
%     Suppose that $\mathcal{L}(\theta^*)=0$ and DDIM Inversion Path follows Eq.\ref{eq:6} then $||(1-m)\odot(z_t^m-\tilde{z}_{t}^{m})||_2^2=||(1-m)\odot(z_t^*-\tilde{z}_{t}^{m})||_2^2$
% \end{theorem}

\begin{proposition}
    % Suppose that $\mathcal{L}(\theta^*)=0$ and DDIM Inversion Path follows Eq.\ref{eq:6}, then $||(1-m)\odot(z_t^m-\tilde{z}_{t}^{m})||_2^2=||(1-m)\odot(z_t^*-\tilde{z}_{t}^{m})||_2^2$ and $E_{z_0}||(1-m)\odot(z_t^*-\tilde{z}_{t}^{m})||_2^2 = 2 \alpha_t {(\sqrt{\frac{1}{\alpha_{t}} - 1} -\sqrt{\frac{1}{\alpha_{t+1}} - 1} ) }^2  \cdot (tr(I)-tr(I^m))$   

        Suppose that $\mathcal{L}(\theta^*)=0$ and DDIM Inversion Path follows Eq.\ref{eq:6} then $||(1-m)\odot(z_t^m-\tilde{z}_{t}^{m})||_2^2=||(1-m)\odot(z_t^*-\tilde{z}_{t}^{m})||_2^2$
\end{proposition}

\begin{proof}
    Due to $z_T^*=z_T^m+m \odot \epsilon$ and $(1-m) \odot z_T^*= z_T^m$
    \label{proof:2}
    \begin{align}
        z_{T-1}^m & = \sqrt{ \alpha_{T-1} \over \alpha_T}z_{T}^m +(1-m) \odot (\sqrt{{ 1 \over \alpha_{T-1} }-1} - \sqrt{{1 \over \alpha_T} -1 })\cdot\epsilon_{\theta}(z_T^{m},T,C_2) \\
           &= (1-m) \odot \sqrt{ \alpha_{T-1} \over \alpha_T}z_{T}^*+(1-m) \odot (\sqrt{{ 1 \over \alpha_{T-1} }-1} - \sqrt{{1 \over \alpha_T} -1 })\cdot\epsilon_{\theta}(z_T^{m},T,C_2) 
    \end{align}
    From Eq.\ref{eq:7}, we denoise $z_{T}^*$ as follows. 
    \begin{align}
        z_{T-1}^*  = \sqrt{ \alpha_{T-1} \over \alpha_T}z_{T}^* + (\sqrt{{ 1 \over \alpha_{T-1} }-1} - \sqrt{{1 \over \alpha_T} -1 })\cdot((1-m)\odot \epsilon_{\theta}(z_T^{m},T,C_2)+m\odot\epsilon_{\theta}(z_T^{*},T,C_1) ) \\
        (1-m)\odot z_{T-1}^*  = \sqrt{ \alpha_{T-1} \over \alpha_T}(1-m) \odot z_{T}^* + (\sqrt{{ 1 \over \alpha_{T-1} }-1} - \sqrt{{1 \over \alpha_T} -1 })\cdot((1-m)\odot \epsilon_{\theta}(z_T^{m},T,C_2))
    \end{align}
    
    Then, $(1-m) \odot z_{T-1}^* =z_{T-1}^m $. For arbitrary $k \leq T$, we also prove following equality.

    \begin{align}
        (1-m) \odot z_{T-k-1}^*  &= \sqrt{ \alpha_{T-k-1} \over \alpha_{T-k}}z_{T-k}^* + (\sqrt{{ 1 \over \alpha_{T-k-1} }-1} - \sqrt{{1 \over \alpha_{T-k}} -1 })\cdot((1-m)\odot \epsilon_{\theta}(z_{T-k}^{m},T-k,C_2))
    \end{align}

    Suppose $t=T-k-1$, then $||(1-m)\odot(z_t^m-\tilde{z}_{t}^{m})||_2^2=||(1-m)\odot(z_t^*-\tilde{z}_{t}^{m})||_2^2$
\\

\end{proof}

\subsection{Details of attention mechanism in Ours}
\label{Appendix: attention}
As mentioned before, U-Net is constructed with self-attention and cross-attention layers. In this section, we elaborate on each mechanism with linear layers. 

Firstly, we mask the cross-attention with a resized mask, $m'$. Spatial feature from the latent $z_t$ is transformed to the Query vector, $Q=\ell_{Q} (\phi(z_t))$, where $\ell_Q$ is linear projection and $\phi$ is spatial feature network of latent. Similarly, we extend key and value vector; $[K_{def},K_{bg}]=[\ell_K(C_{def}),\ell_K(C_{bg})]$ and $[V_{def},V_{bg}]=[\ell_V(C_{def}),\ell_V(C_{bg})]$. From the definition, we define the output of cross-attention with $z_t$ as follows:

\begin{equation}
    Attn^C(z_t,C_{1}) =m' \odot Softmax({QK_{def}^T \over \tau}) V_{def}+(1-m') \odot Softmax({QK_{bg}^T \over \tau}) V_{bg}
\end{equation}

Secondly, we define the self-attention, where we do not directly mask the self-attention. Spatial feature from the masked latent $z_t^m$ is transformed to the Query vector, $\hat{Q}^m=\ell_{\hat{Q}} (\hat{\phi}(z_t^m))$, where $\ell_{\hat{Q}}$ is linear projection and $\hat{\phi}$ is spatial feature network of latent in self-attention layer. Similarly, we extend key and value vector like cross-attention layers, and we define the output of self-attention as follows:

\begin{align}
    Attn^S(z_t^m) \notag & =Softmax({\hat{Q}^m \hat{K}^{mT}\over \tau })\hat{V}^m \\ 
       \notag &=Softmax({\hat{Q}^m\hat{K}^{mT}\over \tau })(1-m') \odot \hat{V}^m \\ 
       &=(1-m')\odot Attn^S(z_t^m) 
\end{align}
\subsection{Details of Qualitative Analysis}

To qualitatively demonstrate performance gain of anomaly detection given our synthetic dataset, we conduct visualizing the loss landscapes given train real anomaly and normal instances. Given trained detection model parameter $\theta_{det}$, the loss landscapes as follows:

\begin{equation}
    g(\alpha,\beta)={1 \over n} \sum_{i=1}^n l(x_i,y_i;\theta_{det}+\alpha d_1+\beta d_2)
\end{equation}

Followed by Anomaly diffusion \cite{hu2024anomalydiffusion}, we utilize the Focal Loss, $l$ with random direction $d_1$,$d_2$ and scaling scalar value $\alpha$, $\beta$.

\subsection{Additional Defect Synthetic Data}
\label{Appendix:additional defect}
\begin{figure}[h!]
    \centering
    \includegraphics[width=\linewidth]{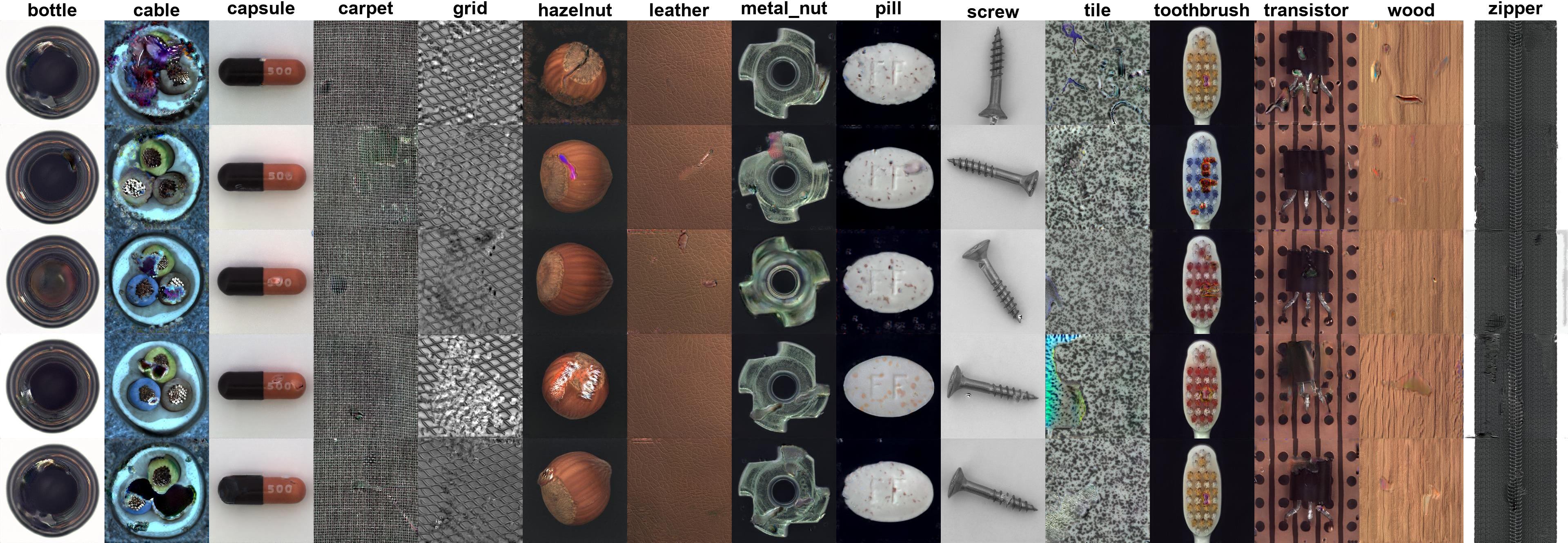}
    \caption{Additional defect generation of DFMGAN in MVTec-Ad}
    \label{fig:dfmgan_gen}
\end{figure}
    
\begin{figure}[h!]
    \centering
    \includegraphics[width=\linewidth]{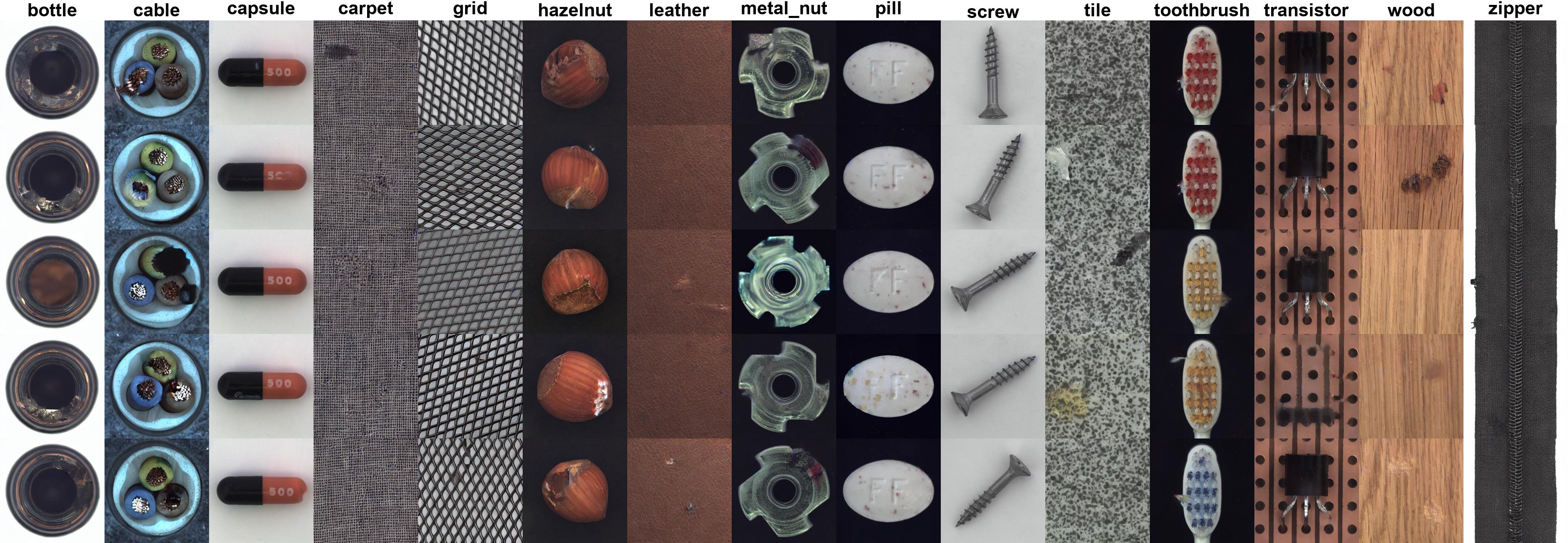}
    \caption{Additional defect generation of Anomalydiffusion in MVTec-Ad}
    \label{fig:ad_gen}
\end{figure}
    
\begin{figure}[h!]
    \centering
    \includegraphics[width=\linewidth]{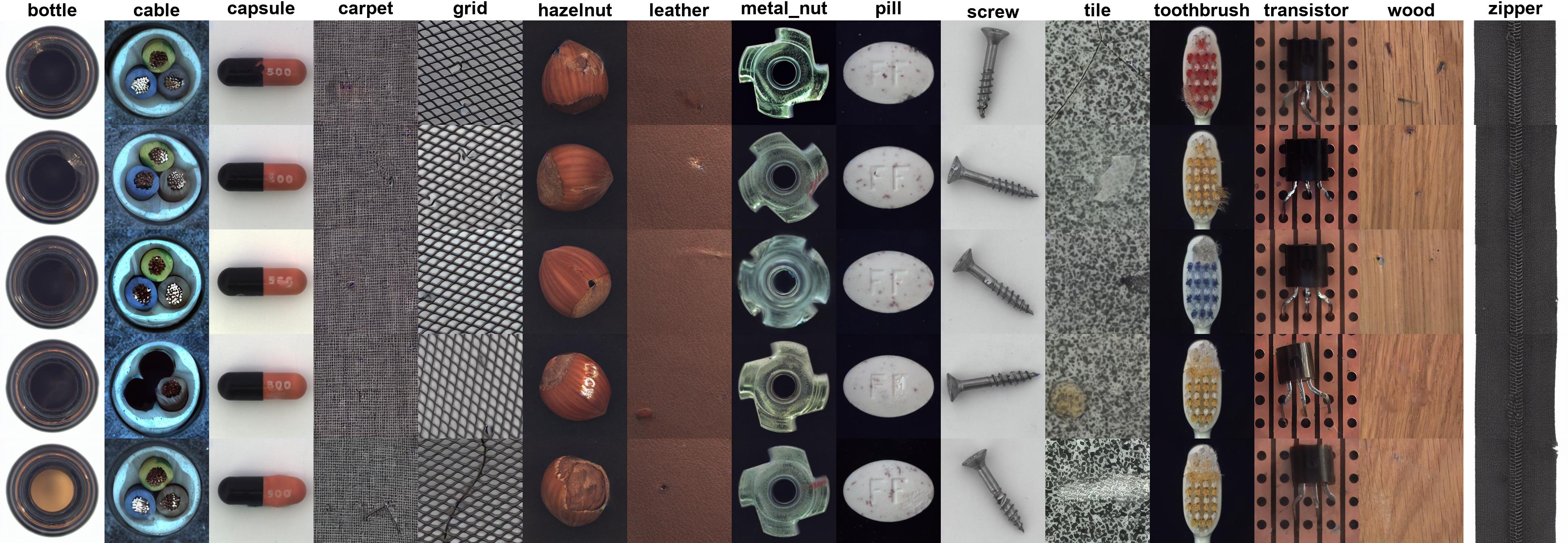}
    \caption{Additional defect generation of Ours in MVTec-Ad}
    \label{fig:ours_gen}
\end{figure}

\begin{figure}[h!]
    \centering
    \includegraphics[width=\linewidth]{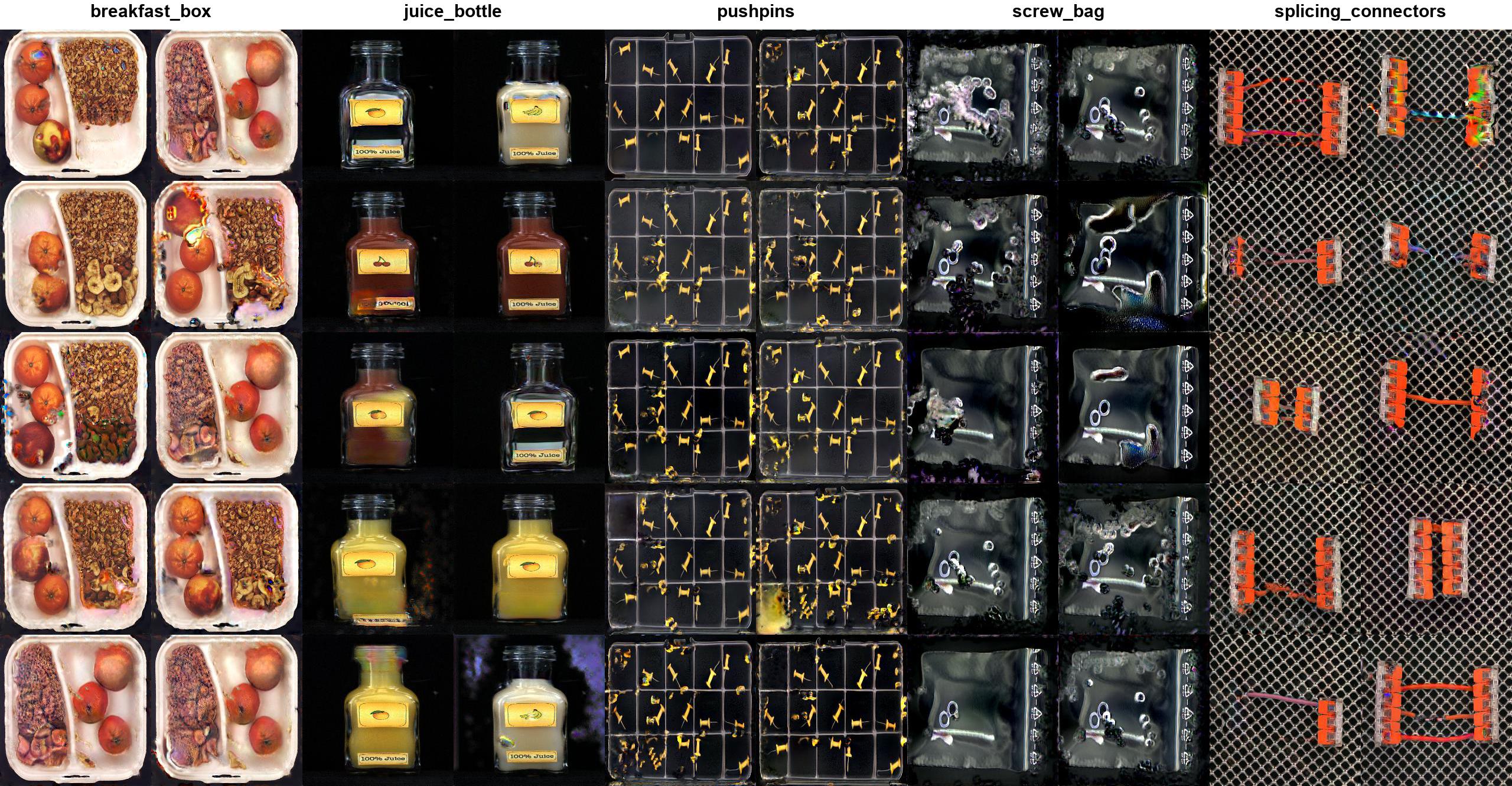}
    \caption{Additional defect generation of DFMGAN in MVTec-Loco}
    \label{fig:dfmgan_gen_loco}
\end{figure}

\begin{figure}[h!]
    \centering
    \includegraphics[width=\linewidth]{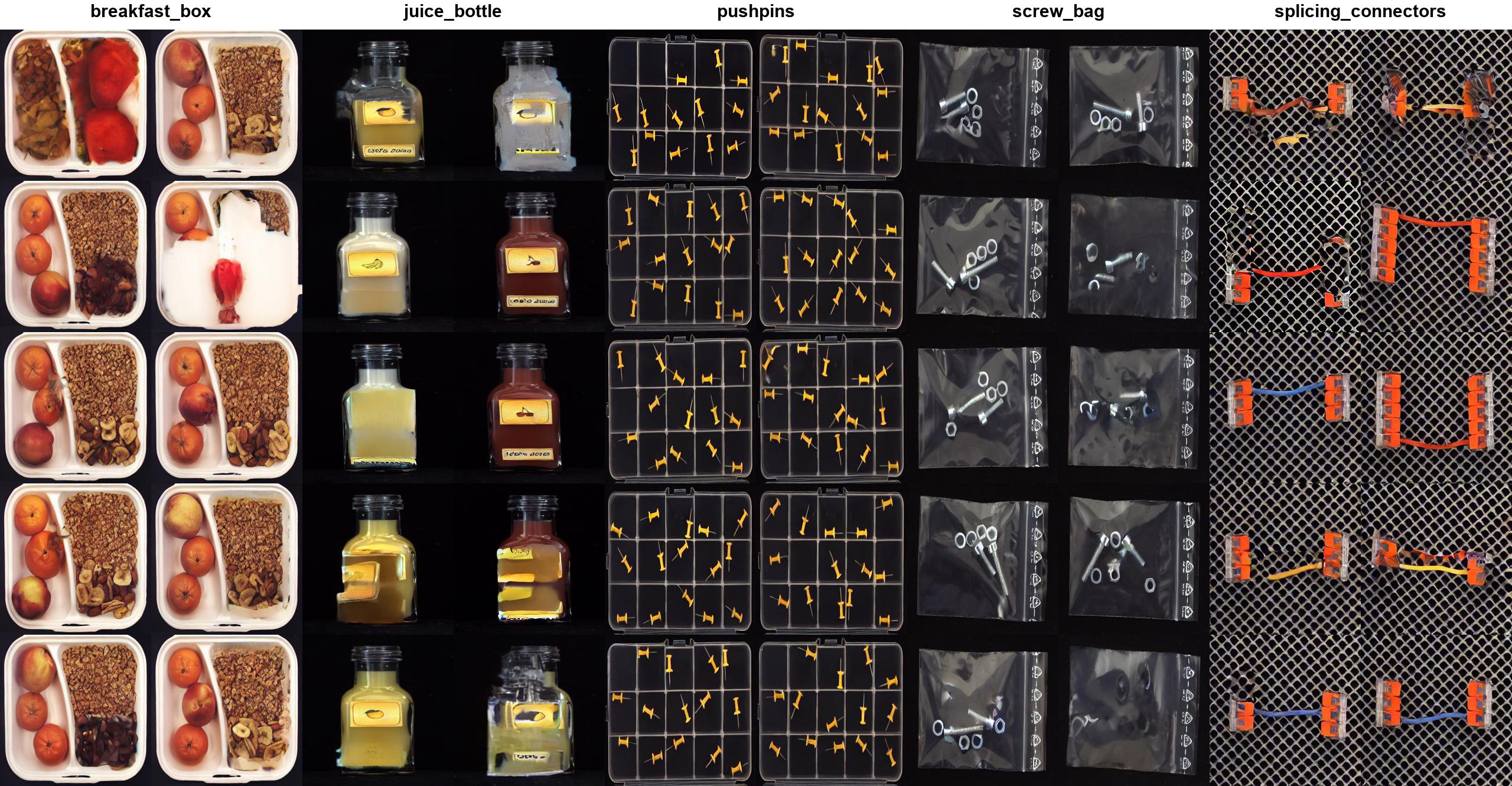}
    \caption{Additional defect generation of Anomalydiffusion in MVTec-Loco}
    \label{fig:ad_gen_loco}
\end{figure}
    
\begin{figure}[h!]
    \centering
    \includegraphics[width=\linewidth]{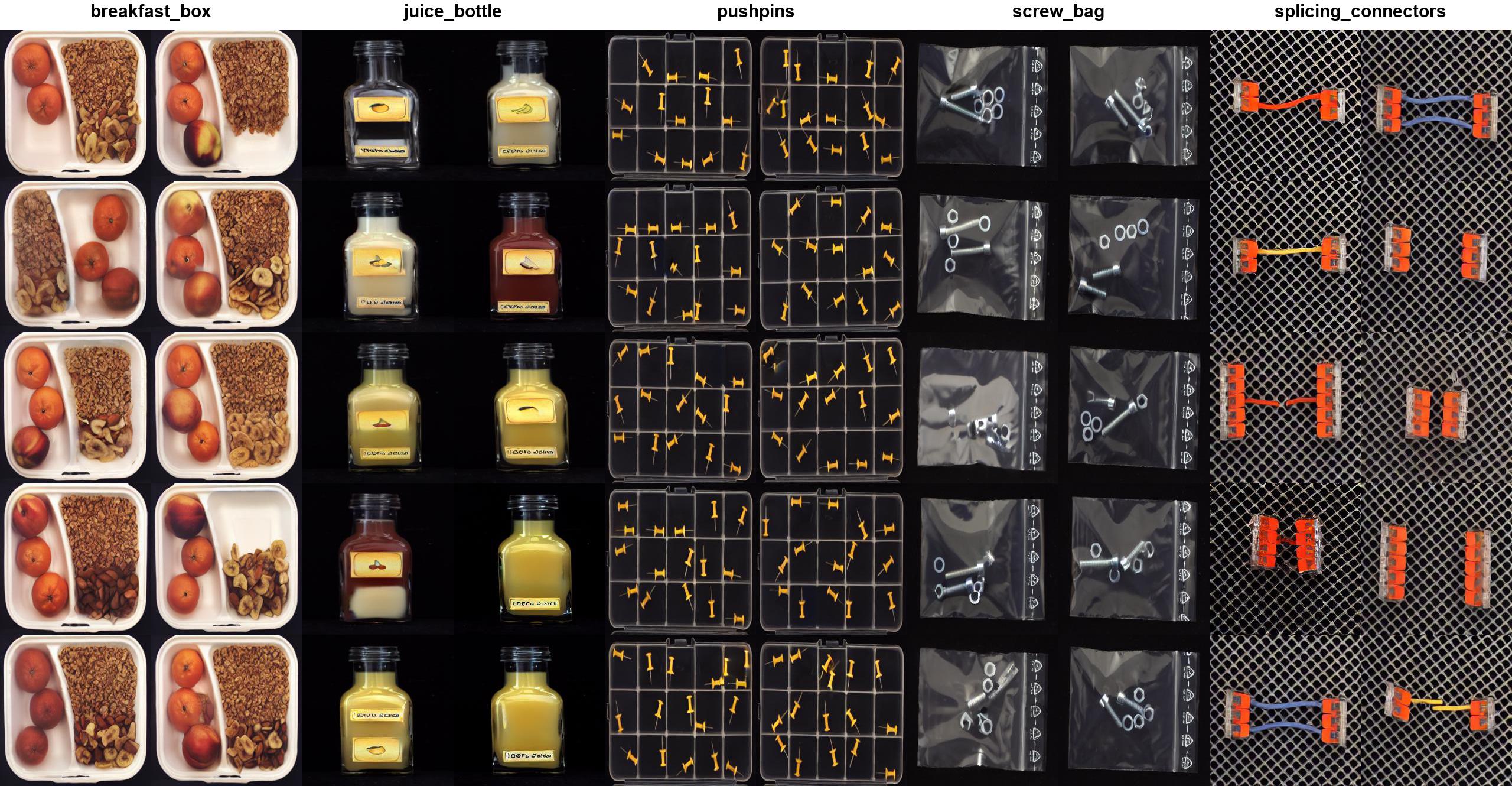}
    \caption{Additional defect generation of Ours in MVTec-Loco}
    \label{fig:ours_gen_loco}
\end{figure}

\end{document}